\documentclass{article}

\usepackage[final,nonatbib]{nips_2018}

\usepackage[utf8]{inputenc} 
\usepackage[T1]{fontenc}    
\usepackage{hyperref}       
\usepackage{url}            
\usepackage{booktabs}       
\usepackage{amsfonts}       
\usepackage{nicefrac}       
\usepackage{microtype}      

\usepackage{amssymb}
\usepackage{latexsym}
\usepackage{nicefrac}
\usepackage{amsmath}
\usepackage{graphicx}
\usepackage[caption=false]{subfig}
\usepackage{lipsum} 

\usepackage{url}
\usepackage{xcolor}
\definecolor{newcolor}{rgb}{.8,.349,.1}
\usepackage[numbers]{natbib}

\title{A Bandit Framework for Optimal Selection of Reinforcement Learning Agents}

\author{
  Andreas Merentitis\thanks{This work was done while the author worked at Zalando Research, Charlottenstrasse 4, 10969 Berlin, Germany} \\
  OLX Berlin Hub\\
  Ella-Trebe-Straße 3, 10557 Berlin, Germany \\
  \texttt{andreas.merentitis@ieee.org} \\
  \And
  Kashif Rasul \\
  Zalando Reasearch \\
  Mühlenstrasse 25, 10243 Berlin, Germany  \\
  \AND
  Roland Vollgraf \\
  Zalando Reasearch \\
  Mühlenstrasse 25, 10243 Berlin, Germany \\
  \And
  Abdul-Saboor Sheikh \\
  Zalando Reasearch \\
  Mühlenstrasse 25, 10243 Berlin, Germany  \\
  \And
  Urs Bergmann \\
  Zalando Reasearch \\
  Mühlenstrasse 25, 10243 Berlin, Germany  \\
  \texttt{urs.bergmann@zalando.de} \\
}

\begin{document}

\maketitle

\begin{abstract}
Deep Reinforcement Learning has been shown to be very successful in complex games, e.g. Atari or Go.  These games have clearly defined rules, and hence allow simulation. 
In many practical applications, however, interactions with the environment are costly and a good simulator of the environment is not available. 
Further, as environments differ by application, the optimal inductive bias (architecture, hyperparameters, etc.) of a reinforcement agent depends on the application.
In this work, we propose a multi-arm bandit framework that selects from a set of different reinforcement learning agents to choose the one with the best inductive bias. To alleviate the problem of sparse rewards, the reinforcement learning agents are augmented with surrogate rewards. This helps the bandit framework to select the best agents early, since these rewards are smoother and less sparse than the environment reward. The bandit has the double objective of maximizing the reward while the agents are learning and selecting the best agent after a finite number of learning steps. 
Our experimental results on standard environments show that the proposed framework is able to consistently select the optimal agent
after a finite number of steps, while collecting more cumulative reward compared to selecting a sub-optimal architecture or uniformly alternating between different agents.
\end{abstract}

\section{Introduction}

Reinforcement learning has been successfully used in numerous application domains, from robot control \citep{DBLP:conf/cdc/GrondmanBB12}, \citep{DBLP:journals/tsmc/AdamBB12} 
to investment management and board games. Recent advances in the field such as \citep{DBLP:journals/tsmc/GrondmanBLB12}, \citep{icml2014c1_silver14}, 
\citep{DBLP:journals/corr/ODonoghueMKM16}, \citep{DBLP:journals/corr/NachumNXS17} have enabled state of the art reinforcement learning systems to steer autonomous cars, compete with the human 
champions in games such as Chess and Go, as well as achieve superhuman performance in many of the old classic Atari games \citep{DBLP:journals/corr/MnihBMGLHSK16}, 
\citep{DBLP:journals/corr/SchulmanLMJA15}. Reinforcement learning in its most basic form includes  an agent that interacts with its environment by taking actions. As a result of these actions 
there is a change of state and the agent might also receive a reward. The goal of the system is to learn an optimal policy, i.e. learning how to act so that it can maximize its cumulative reward.

In many practical reinforcement learning settings, we do not have access to a perfect simulator of the environment and interacting with the real environment is costly. 
In addition, we want to gain good rewards even early on, while the agent is still learning the environment dynamics. Model-based reinforcement learning 
assumes a priori a certain model class for the environment dynamics which can be beneficial (e.g., reducing the interactions with the environment needed to learn) 
if the model is close to the true dynamics, but detrimental (biasing the agent's actions) if it is not. While this restriction is removed in model-free reinforcement learning, 
the architecture of the agent is also encoding prior beliefs about the environment and its transition dynamics (for example regarding the level of noise, 
assuming full observability or not, etc.). At the same time, if interactions with the environment are costly, learning multiple agents that represent
different beliefs about the environment dynamics quickly becomes expensive. In this paper we are trying to systematically address two questions: ``how can we maximize 
rewards while learning?'' and ``how to select the optimal agent configuration (and consequently policy) after a finite number of learning steps?''

In order to address these questions we consider a bandit framework on top of a set of reinforcement learning agents. Our goal then is to take actions (pull levers in bandit terminology, which corresponds to selecting one of several RL agents in this framework) 
so as to minimize the regret, i.e. the difference in reward between the sequence of selections that we have made and the sequence of selections performed by an oracle that always makes the best choice.

In addition to the bandit framework, we also consider augmenting the environment reward with an agent specific surrogate reward, that captures the inherent ability of the 
agent to understand the dynamics of the environment. This surrogate reward is inspired by the Variational Information Maximizing Exploration idea \citep{DBLP:journals/corr/HouthooftCDSTA16}, 
where a similar metric, which captures the surprise of the agent regarding the environment dynamics, is used to favor exploration of the state space (based on equation \ref{eq:4}). However, in this 
work the surrogate reward is used to facilitate early selection between alternative agent architectures and gather more reward while learning.

The rest of the paper is structured as follows. Some background information on bandits and reinforcement learning, as well as an outline of the state of the art, is provided in
Section 2. The proposed framework for optimizing the reward while learning and facilitating the selection of an optimal agent after a finite number of steps is
elaborated in Section 3. This part introduces also the idea of an augmented surrogate reward that is both smoother and less sparse than the actual reward of the environment
and discusses how such a reward can be derived from agents that employ Bayesian Neural Networks. Experimental results for several environments including some classic 
locomotion reinforcement learning problems as well as a few more modern Atari environments are given in Section 4. Finally section 5 provides some concluding remarks.

\section{Background}
In reinforcement learning we assume a Markov Decision Process (MDP), defined by its state space X, its action space U, its transition probability function 
$f : X \times U \times X \rightarrow [0, \infty)$, and its reward function $\rho : X \times U \times X \rightarrow \mathbb{R}$. At every discrete time step $t$, given the state $x_t$ , 
the agent selects an action $u_{t}$ according to a policy $h : X \rightarrow U$. The probability that the next state $x_{t+1}$ falls in a region $X_{t+1} \subset X$ of the state space 
can then be expressed by $\int_{X_{t+1}} f(x_{t} , u_{t} , x' ) dx' $. For any $x$ and $u$, $f (x, u, \cdot)$ is assumed to define a valid probability density of the argument “$\cdot$”. After the 
transition to $x_{t+1}$, a (possibly stochastic) reward $r_{t+1}$ is derived according to the reward function $\rho$ : $r_{t+1} = \rho (x_{t} , u_{t} , x_{t+1} )$. The reward depends on the current
and next state, as well as the current action -- but even conditioned on these it is generally drawn from a probability distribution (for non deterministic MPDs). For deterministic MDPs, the transition 
probability function $f$ can be replaced by the transition function, $f : X \times U \rightarrow X$, and the reward can be fully determined by the current state and action: $r_{t+1} = 
\overline {\rho} (x_{t} , u_{t})$, $\overline {\rho} : X \times U \rightarrow \mathbb{R}$.

The goal is to find an optimal policy $h^*$ that maximizes the expected return for every initial state $x_{0}$. In practice, the accumulated return must be maximized using 
only feedback about the immediate, one-step reward. Every policy $h$ is characterized by its state-action value function (Q-function), $Q^h : X \times U \rightarrow \mathbb{R}$, 
which gives the return when starting in a given state, applying a given action, and following $h$ from that point onward. For any $h$, $Q^h$ is unique and can be found by solving 
the Bellman expectation equation, while the optimal Q-function is defined as $Q^* (x, u) = \max_{h} Q^h (x, u)$, and satisfies the Bellman optimality equation.

Multi-arm bandits can be considered a special case of reinforcement learning, in which the Markov decision process only has one state and the actions cannot change the environment,
but the MDP is not deterministic in the sense that the rewards are drawn from a probability distribution. Formally, a bandit seeks to minimize the difference (regret) between the actions 
of an oracle that always selects the best action and the choices we make using the bandit strategy. 

Different types of bandit strategies are possible, from simple $\epsilon$-greedy, to more sophisticated techniques like SoftMax (probability matching bandit), UCB1 (a bandit 
based on the optimism in the face of uncertainty principle) and EXP3 (adversarial bandit). 
For a more thorough discussion of bandit algorithms please refer to dedicated survey papers like \citep{Vermorel2005multi}.

\section{Multi-arm bandit framework}
In many real world reinforcement learning settings, a good simulator of the environment is not available. Further, interactions with the environment are costly, either
directly (e.g., having an immediate negative monetary impact in case of incorrect decisions) or indirectly (e.g., negatively affecting customer perception).
In such cases, we are particularly motivated to get high rewards even early on while the agent is still learning the environment dynamics. In model-free reinforcement learning, 
the agents architecture reflects our prior beliefs about the environment and its transition dynamics (for example regarding the level of noise, observability assumptions, etc.). 
Typically though, we are not certain in advance which of these prior beliefs are true and to what extent, therefore trying multiple agent architectures might be beneficial. 
However, if interactions with the environment are costly, learning multiple agents that represent different beliefs about the environment dynamics quickly becomes expensive. 
The goal of the multi-arm bandit framework is to maximize rewards while learning, but to also provide a systematic method for selecting an optimal agent  
from a set of candidates, after a finite number of learning steps.

The bandit framework is supplemented by agent specific surrogate rewards that capture the inherent ability of the agents to model the environment. These surrogate rewards are inspired by the Variational Information Maximizing Exploration concept, where a metric capturing the surprise of an agent regarding the environment dynamics is used to promote exploration of the state space. The key idea behind introducing a surrogate reward metric is that the true environment rewards can be very sparse and noisy, especially early on while the agents are still learning \citep{DBLP:journals/corr/HouthooftCDSTA16}. Since we are interested in putting the bulk of our actions on agents that are at least promising in their ability to learn the dynamics of the environment (compared for the same amount of exploration), this less sparse reward is a good way to achieve that goal. One of the key contributions of our work over standard VIME is that we use the surrogate reward as a way to guide the selection of different reinforcement learning algorithms by the bandits. More specifically, the properties of this surrogate reward being smoother and capturing how ``surprised'' the agents are from the responses of the environment are also very beneficial, both with respect to focusing early onto the most promising agents, as well as on being able to select the best agent after a relatively small number of steps. 

It has been shown in \citep{DBLP:journals/corr/abs-1103-5708} that it is 
beneficial for agents to take actions that maximize the reduction in uncertainty about the environment dynamics. This can be formalized as taking a sequence of actions $a_t$ that maximize the sum of reductions in entropy. With the history of the agents up until time step $t$ as $\xi_{t} = \{s_1 , a_1 , \dots , s_t \}$, we can write the sum of entropy reductions as

\begin{equation}
        \sum_{t} \left( H\left(\bold\mathbf{\Theta}| \xi_{t}, a_t \right) - H \left(\bold\mathbf{\Theta} | S_{t+1}, \xi_{t}, a_t \right) \right). \label{eq:1}
\end{equation}
As indicated in \citep{DBLP:journals/corr/HouthooftCDSTA16}, according to information theory, the individual terms express the mutual information between the next state distribution $S_{t+1}$ 
and the model parameter distribution $\bold\mathbf{\Theta}$, namely $I(S_{t+1}; \bold\mathbf{\Theta}|\xi_{t},a_{t})$. This implies that an agent is encouraged to take actions that are as informative as possible 
regarding the environment transition dynamics. This mutual information can be written as:

\begin{equation}
       I \left (S_{t+1};\bold\mathbf{\Theta}|\xi_{t},a_{t} \right) = \mathbb{E}_{s_{t+1} \sim P \left(\cdot|\xi_{t},a_{t} \right)} [D_{KL}[p \left(\theta|\xi_{t},a_t,s_{t+1}\right) || p \left(\theta|\xi_{t} \right)]], \label{eq:2}
\end{equation}
where the KL divergence term is expressing the difference between the new and the old beliefs of the agent regarding the environment dynamics, and the expectation is with respect
to all possible next states according to the true dynamics. Under these assumptions the above formulation can also be interpreted as information gain \citep{DBLP:journals/corr/HouthooftCDSTA16}.

Formally, since calculating the posterior $p(\theta|D)$ for a dataset $D$ is not feasible, we follow VIME and approximate it through an alternative distribution $q(\theta;\phi)$, parametrized by $\phi$. In this setting we seek to minimize $D_{KL}$ through maximization of the variational lower bound $L[q(\theta ; \phi),D]$. The latter is formulated as:

\begin{equation}
        L[q(\theta ; \phi),D] =  \mathbb{E}_{\theta \sim q(\cdot ; \phi)} [\log p(D|\theta)] - D_{KL}[q(\theta ; \phi) || p(\theta)]. \label{eq:3}
\end{equation}

The information gain term can then can be expressed as:

\begin{equation}
       I(S_{t+1}; \bold\mathbf{\Theta}|\xi_{t},a_{t}) = D_{KL} [q(\theta ; \phi_{t+1}) || q(\theta ; \phi)], \label{eq:4}
\end{equation}
where $\phi_{t+1}$ represents the updated and $\phi_{t}$ the old parameters of the agent's belief regarding the environment dynamics. Since Bayesian Neural Networks are expressive parameterized models that can maintain a distribution over their weights, they are excellent candidates for realizing reinforcement learning agents which can directly provide an estimate of the information gain via equation \ref{eq:4}. 

In our setting we are interested in keeping a metric that captures how certain a given agent is about the environment dynamics after a fixed amount of exploration, as opposed to a metric that only favors exploration like in the original VIME paper \citep{DBLP:journals/corr/HouthooftCDSTA16}. The reason for this deviation is that we are on the one hand interested to explore the environment in order to learn a good policy early (similar to standard VIME), but on the other hand we want to be able to compare between the different reinforcement learning algorithms and architectures, using their ``surprise'' for a fixed amount of environment iterations as a metric. Therefore we consider a moving average over the normalized inverse formulation of the previous term for a specified number of environment interactions (since less ``surprise'' means better understanding of the environment, if the number of interactions is fixed).

\section{Experiments}

In order to validate the idea of using a bandit framework on top of a collection of reinforcement learning agents, we use a number of well known environments that include
some traditional reinforcement learning tasks (e.g., Acrobot, Cartpole), some basic locomotion tasks (e.g., Mountain Car, Lunar Lander), as well as some classic games such as KungFu, Ms. Pacman, Space Invaders, Zaxxon, Seaquest, and Atlantis from the Atari collection \citep{DBLP:journals/corr/BrockmanCPSSTZ16}. 

We analyze different types of bandit algorithms, each of which has to select among a set of different architectures of reinforcement learning agents: vanilla VIME for the simple locomotion tasks (with different sets of hyperparameters), and VIME combined with Deep Q-Network (DQN) or Asynchronous Actor-Critic Agents (A3C) for the Atari environments (again with several hyperparameter options, one captured in each agent). In addition, we also compare the results of the best agents in our framework with literature results of DQN agents from \citep{mnih-dqn-2015}, as well as with the more recent massively parallelized form of the same framework, Gorila \citep{DBLP:journals/corr/NairSBAFMPSBPLM15}.

\begin{figure}[h!]
   \begin{tabular}{ll}
    	\hspace{-0.25cm} \includegraphics[width=0.5\columnwidth]{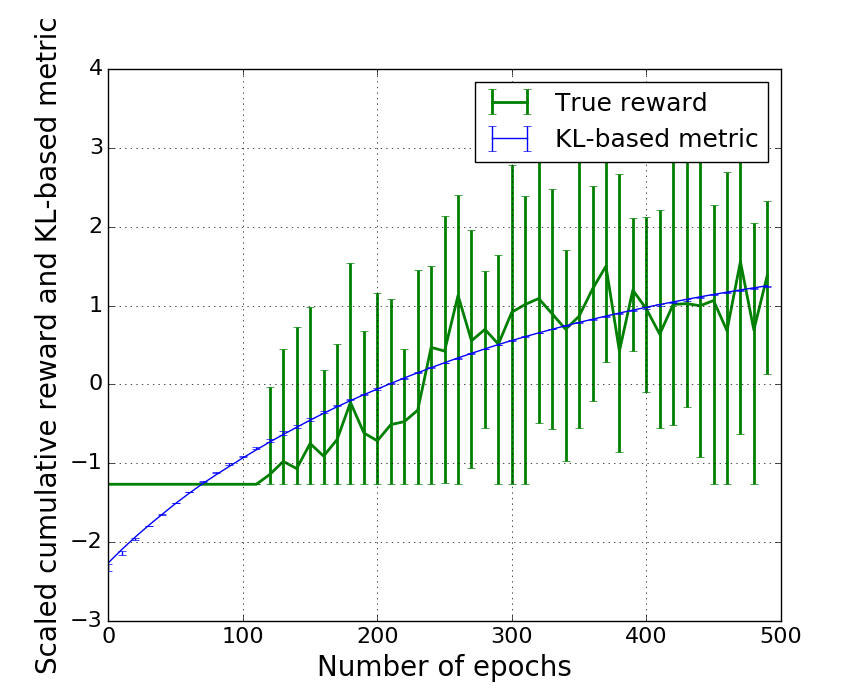} &
			\hspace{-0.25cm} \includegraphics[width=0.5\columnwidth]{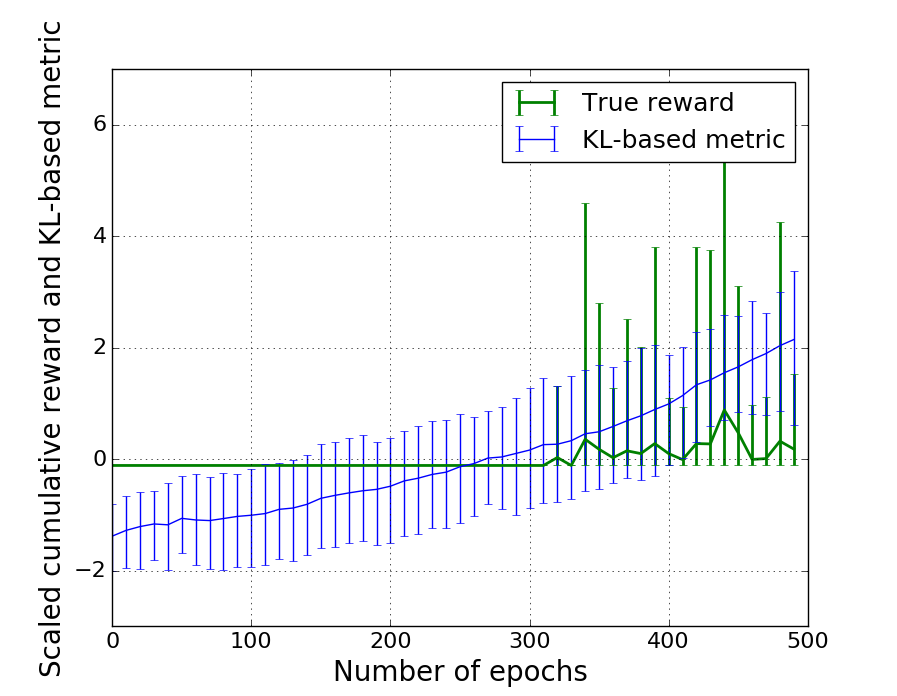} \\
	 \end{tabular}
   \vspace{-0.4cm}\caption{\em Correlation between environment and information gain rewards for a good (left) and a suboptimal agent (right) for Mountain Car environment.}
\label{fig:correlation_check}
\end{figure}

Before analyzing the behaviour of bandits on top of reinforcement learning agents, we investigate single agents. The first step is to show that the surrogate reward (consisting of the true reward and the information gain) is promising for early selection of the reinforcement learning agents. In this direction we select the part of the reward that captures how well the agent is learning the environment dynamics, the information gain, and we plot it
(rescaled) together with the true environment rewards (Figure ~\ref{fig:correlation_check}), for two different agents (one that is particularly good for the task, one that is able to solve it but is suboptimal). Both plots are the averages of 50 instantiations of the environment.

It is clear from Figure ~\ref{fig:correlation_check} that the true rewards of the environment (presented with green) and the information gain rewards (only the part that captures the certainty of 
the agent about the environment dynamics, presented with blue) are highly correlated, both for the good agent as well as for the suboptimal one. In addition, it is apparent that the 
information gain reward is both less sparse, taking informative values much earlier than the true reward (especially for the suboptimal agent) as well as smoother across the 
different instantiations of the environment. A good question then is why do we need the surrogate reward to be composed of two parts, one that captures the agent's certainty
about the environment and another that captures directly the true reward? This is necessary because later on the bandit will need to make increasingly more difficult decisions between agents
that are all good and similar on their ability to model the environment and in this case capturing directly the true reward is necessary to differentiate between them.

\begin{figure}[h!]

\subfloat[Cumulative surrogate reward for the Cartpole and Mountain Car environments (scaled).]{%
\begin{tabular}{ll}
    	\hspace{-0.25cm} \includegraphics[width=0.5\columnwidth]{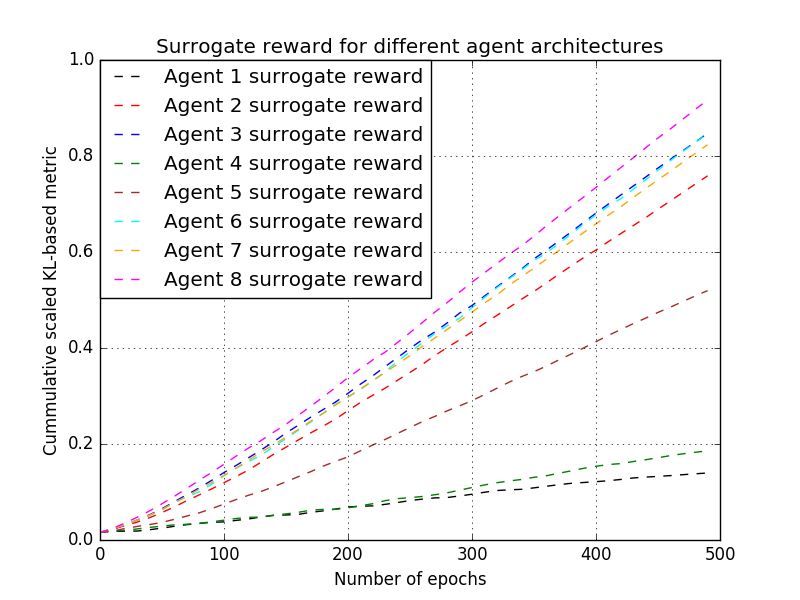} &
			\hspace{-0.25cm} \includegraphics[width=0.5\columnwidth]{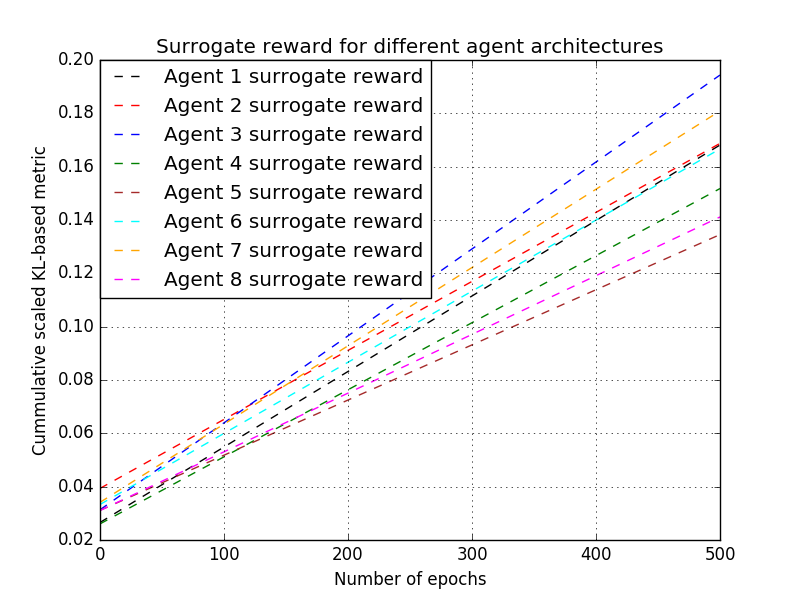} \\
	 \end{tabular}
}

\subfloat[Cumulative true reward for the Cartpole and Mountain Car environments (scaled).]{%
\begin{tabular}{ll}
    	\hspace{-0.25cm} \includegraphics[width=0.5\columnwidth]{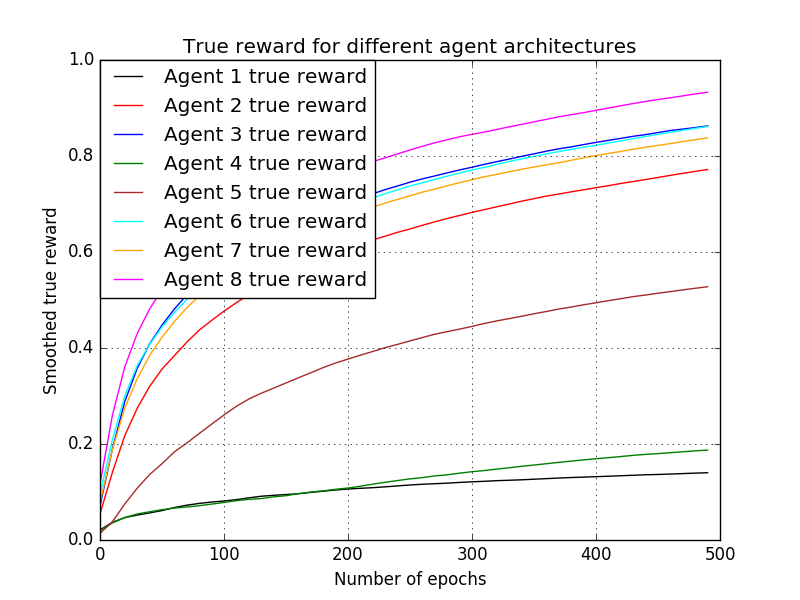} &
			\hspace{-0.25cm} \includegraphics[width=0.5\columnwidth]{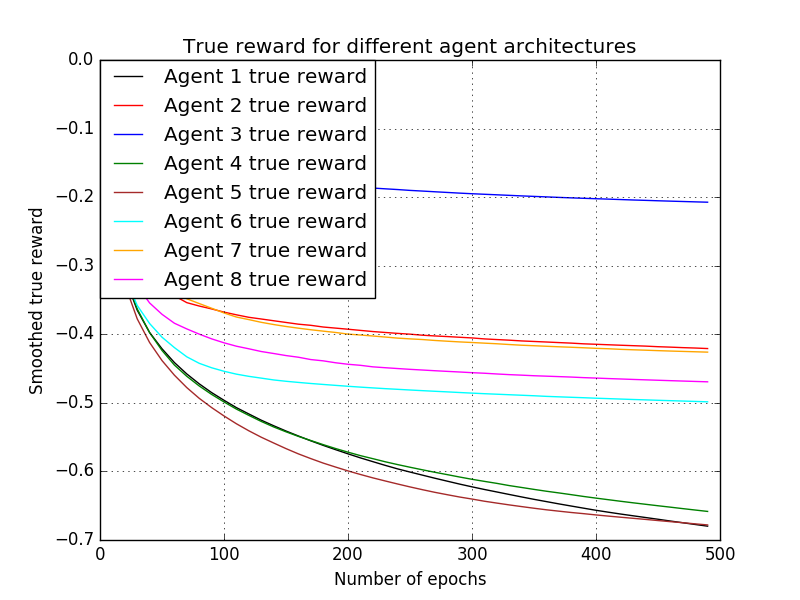} \\
	 \end{tabular}
}

\caption{Comparison of the cumulative surrogate and true rewards for the Cartpole and Mountain Car environments for different reinforcement learning agents. The relative order of the agents is similar between the first and second rows, pointing that the surrogate reward can be used to augment the true reward, especially early on when the latter is sparse and noisy.}
\label{fig:cart_mc_rewards}
\end{figure}

The next step is to show that the surrogate reward allows focusing on the agents that are indeed the most promising ones in terms of their ability to collect reward early on, but also in terms of their ability to learn the environment dynamics so that collection of reward is also maximized in the long run. In order to show this property, we investigate the relation between the full surrogate reward and the true reward, especially for good agent architectures. Again we use the same setup as before, i.e.
the graphs show results that are averages of 50 environment instantiations. In order to interpret the result one has to keep in mind that the surrogate reward is expressing a 
scaled version of the certainty of the agent about the environment dynamics (it is always increasing), while the raw environment rewards can be negative, either in the early stage of
learning only or always. In all cases, the best agents are the ones that have the highest values on Figure ~\ref{fig:cart_mc_rewards}. 
We can see that the three best agents in the figures with the doted lines (the surrogate rewards) are 
also the best ones in terms of the actual environment reward (solid lines) for both cases (note that for some environments highest is the largest positive value, and for some others it is the least negative one). 

Having established that the surrogate reward is correlated with the environment reward and also has the desirable properties of being smoother and less sparse,
we proceed to validate the core part of the bandit framework. Here we seek to answer the three questions: 1) how do the different bandits compare to each other, 2) whether a bandit framework is able to assign most of the actions on promising agents, as well as 3) if a bandit can select the best agent after a finite number of steps. For comparison we add an oracle that always selects the best agent, as well as a baseline that picks the worst agent.

\begin{figure}[h!]
   \begin{tabular}{ll}
    	\hspace{-0.25cm} \includegraphics[width=0.5\columnwidth]{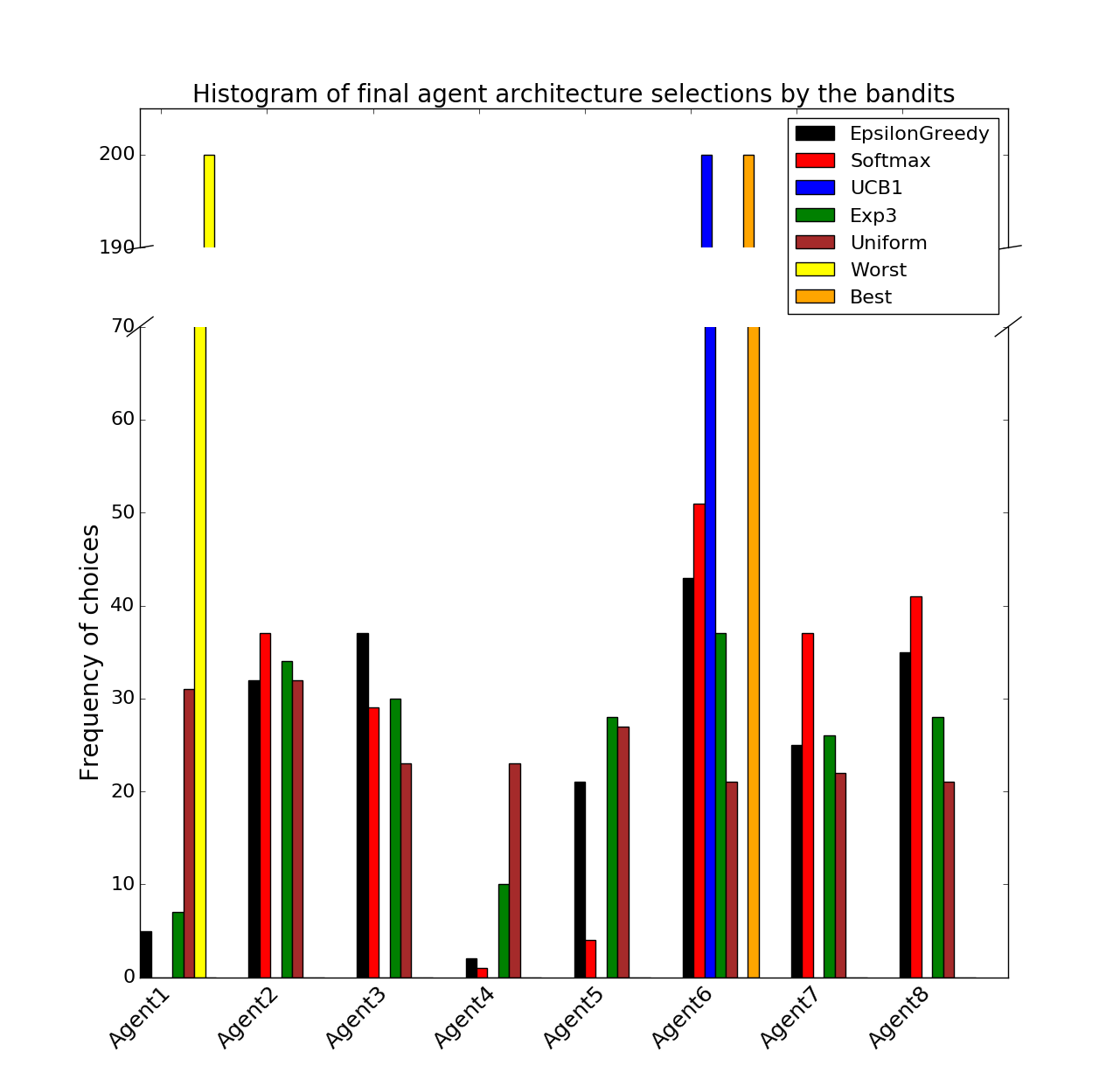} &
			\hspace{-0.25cm} \includegraphics[width=0.5\columnwidth]{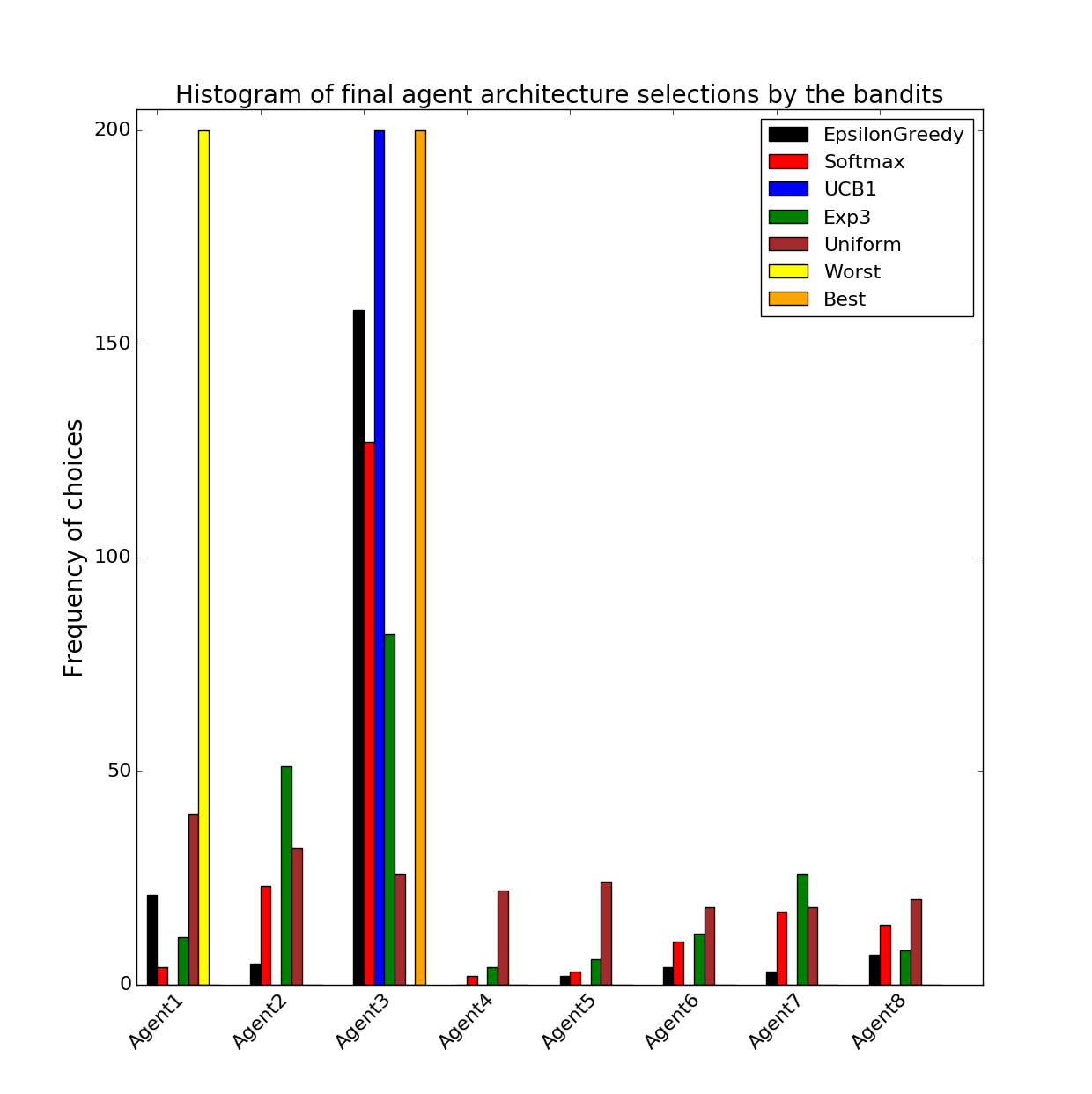} \\
	 \end{tabular}
   \vspace{-0.4cm}\caption{\em Frequency of selections of the different agents for Lunar Lander (left) and Mountain Car (right) at the end of training for the different bandit algorithms. The UCB
   algorithm matches the oracle (Best) in selecting the best agent after training is complete.}
\label{fig:lunar_mc_bar}
\end{figure}

The basic setup of the framework is that we define a window of iterations (typically 10 for locomotion environments and 20 for Atari) after which the bandit reviews the choice of reinforcement learning agent, and can select one out of several alternatives. The selected agent is then allowed to interact with the environment, collecting rewards in the process. The agents tested are vanilla VIME for the simple locomotion tasks (with different sets of hyperparameters taken from the literature as well as slight modifications with increased or decreased layer sizes, one more or one less layers, etc), and VIME combined with Deep Q-Network (DQN) or Asynchronous Actor-Critic Agents (A3C) for the Atari environments again based on literature or slightly modified variations. The bandits we are using for selection between the agents are Epsilon Greedy, Softmax, UCB1 and Exp3, while uniform, worst and best (oracle) are added as reference points.    

First, let's look at the selection behaviour of the bandits after learning.
From Figure ~\ref{fig:lunar_mc_bar} we can see that the UCB1 algorithm was able to pick the best agent architecture for all 200 instantiations of the environments , followed closely by Epsilon Greedy in the case of Mountain Car. However, for Lunar Lander the gap 
between the number of correct picks of the UCB1 algorithm and the next best one (SoftMax bandit in that case) is quite large. This performance gap can be reduced if we select better hyper 
parameters for the non-UCB algorithms, but that is not easy in real world settings, where a simulator of the environment is not available. A possible reason that algorithms such as SoftMax are not doing very well for Lunar Lander is the large number of different possible initial conditions of the game (different starting points of the space shuttle that has to be landed) which makes estimations of reliable expectations of the reward probability associated with each bandit action challenging. For Epsilon Greedy, although it does not calculate posterior probabilities, the large number of combinations makes random exploration less effective as well.   

\begin{figure}[h!]
   \begin{center}
   \includegraphics[width=0.65\columnwidth]{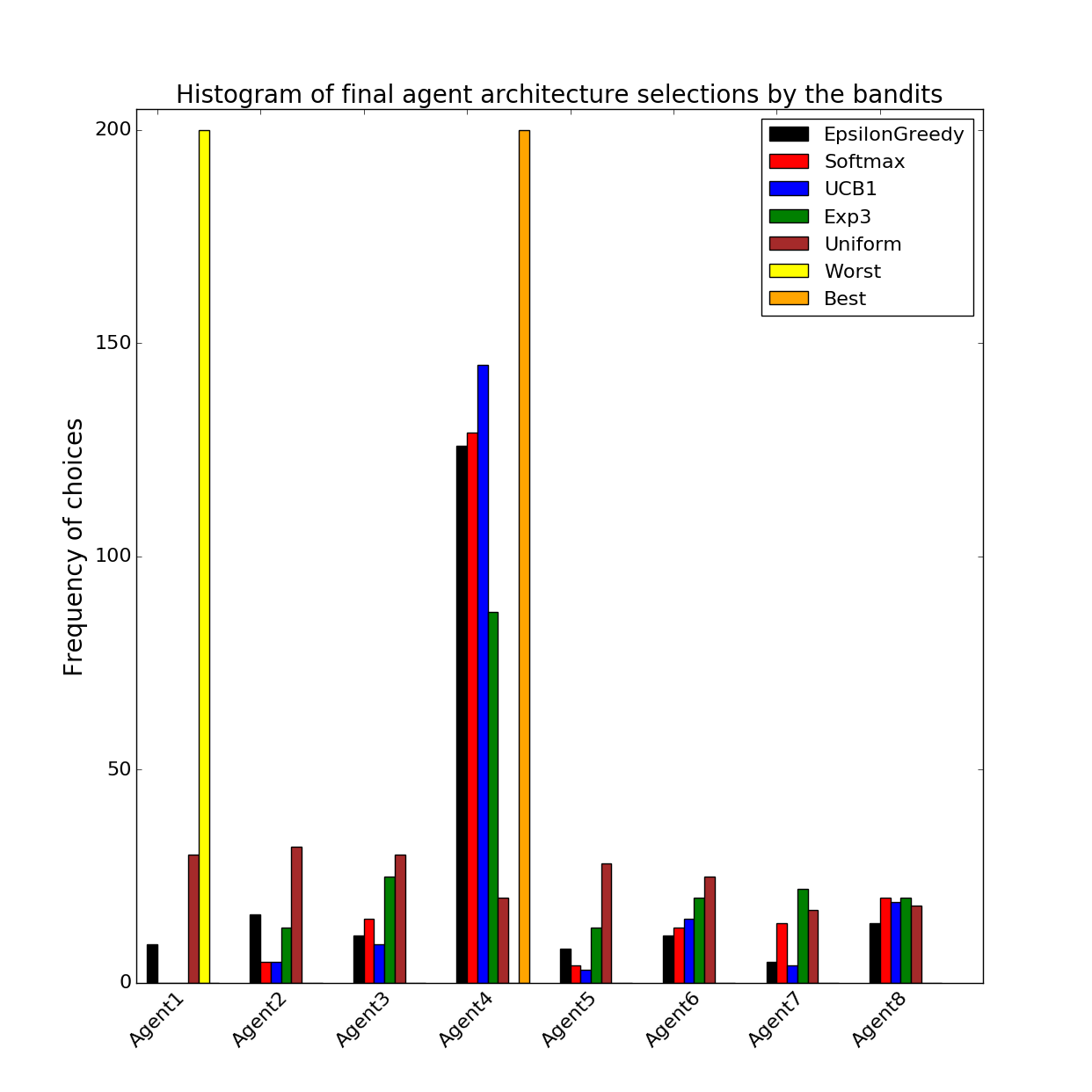}
	 \end{center} 
   \vspace{-0.4cm}\caption{\em Frequency of selections of the different agents for SpaceInvaders at the end of the training period for the different bandit algorithms. In this setting the variance is high and no bandit algorithm can perfectly match the oracle (Best) in the specified number of rounds.}
\label{fig:space_inv_bar}
\end{figure}

For Atari games (Space Invaders is shown as a representative example, Figure ~\ref{fig:space_inv_bar}) one observation is that all bandit algorithms have a somewhat tougher job of selecting 
reliably the best agent. To some extent this is because two or more agents are typically close in terms of performance, but also because the variance of of the agent rewards over
different instantiations of the environment can be significant, even for a well trained agent in some of these games. In fact it is not uncommon for one agent to have the 
potential for high reward, while another one achieves more reliably a certain level of lower, yet relatively good, reward. In such scenarios we might either select algorithms that best capture our preference with respect to the variance of the rewards or allow for a diversification strategy by keeping all the agents that are not strictly inferior to some other agent.

\begin{figure}[h!]
   \begin{tabular}{ll}
    	\hspace{-0.25cm} \includegraphics[width=0.46\columnwidth]{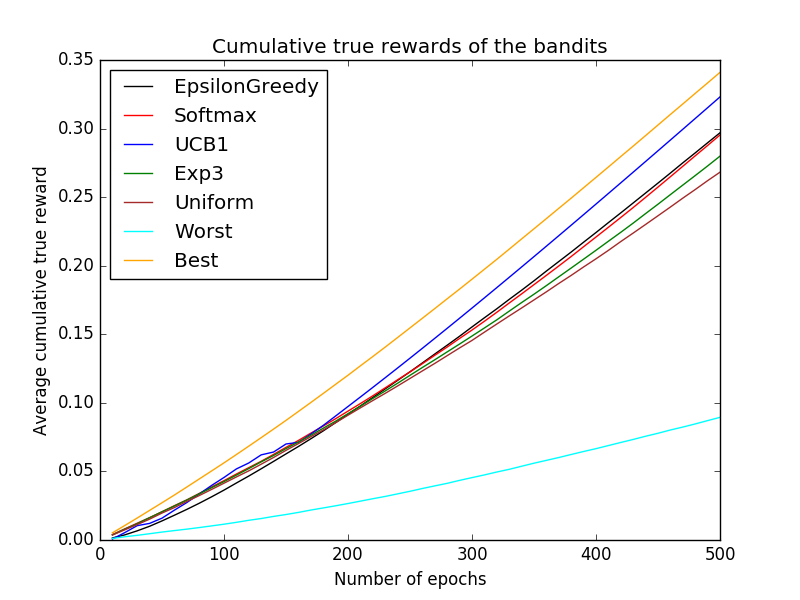} &
			\hspace{-0.25cm} \includegraphics[width=0.55\columnwidth]{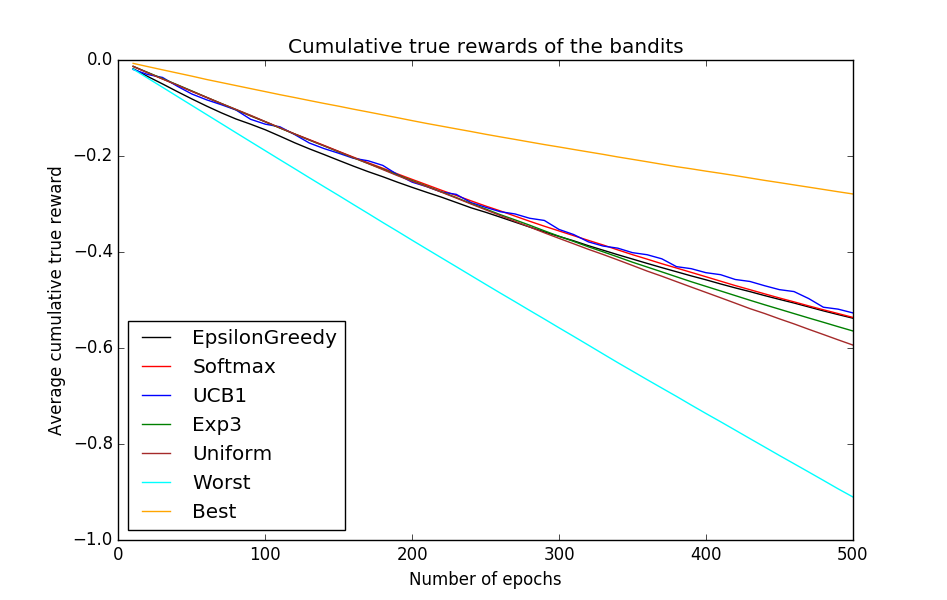} \\
	 \end{tabular}
   \vspace{-0.4cm}\caption{\em Cumulative true reward for the Lunar Lander and Mountain Car environments (scaled).}
\label{fig:lunar_mc_algos}
\end{figure}

\begin{figure}[h!]
   \begin{center}
   \includegraphics[width=0.55\columnwidth]{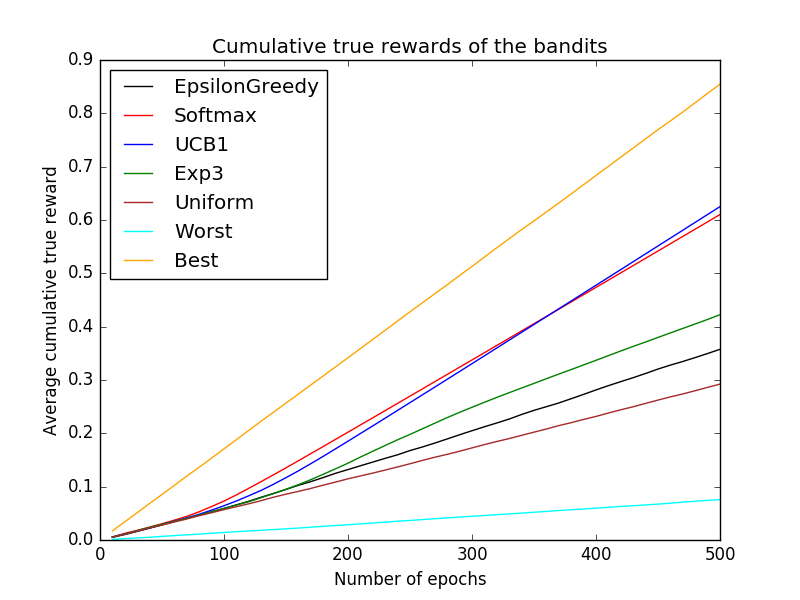}
	 \end{center} 
   \vspace{-0.4cm}\caption{\em Cumulative true reward for the SpaceInvaders environment (scaled).}
\label{fig:space_inv_algos}
\end{figure}

The remaining question to be answered is whether the bandit framework is efficient in terms of collecting reward while learning. For this reason we will compare the cumulative
rewards collected by the different bandit algorithms against 3 baselines: 1) a strategy of uniform alternation between the agents, 2) selecting the worst agent and stay with it, and 3)
the oracle case of always selecting the best agent (Figures ~\ref{fig:lunar_mc_algos} and ~\ref{fig:space_inv_algos}). These 3 baselines will show us different extreme 
cases that are valuable for comparisons. The uniform strategy shows us if there is any benefit from the bandit at all, compared to the case of naively alternating between agents. 
The baseline of picking the worst bandit and staying with it shows if the alternation between agents is justified compared to exploring deeply the environment with a sub-optimal RL agent. In this approach we have to continue with an agent that does not map well to the problem but we do not waste any time in exploring other agents, so all the effort is focused. Unfortunately, in many interesting cases the worst agent fails completely to solve the problem at hand, therefore some exploration is necessary. Finally, the oracle also does not spend time exploring so it provides an estimation of the performance gap between the bandit framework and the ideal but unrealistic case. 

\begin{table*}[!t]
  \caption{\label{tab1} Score for DQN, Gorila, pro human gamer, and the agent selected from the bandit}
  \centering
  \begin{tabular}{|p{2.5cm}|p{2.5cm}|l|p{2.5cm}|p{2.5cm}|}
  \hline
   Games & DQN Score & Gorila Score & Human Pro Score & Best Agent Score\\
  \hline
    Atlantis            & 85641 $\pm$ 17600 & 100069.16 & 29028 & 217810 $\pm$ 7256.4 \\
    Kung-Fu Master      & 23270 $\pm$  5955 & 27543.33  & 22736 & 29860  $\pm$  6793.1 \\
    Ms. Pacman          &  2311 $\pm$   525 & 3233.50   & 15693 & 5708.0 $\pm$  860.1 \\
    Seaquest            &  5286 $\pm$  1310 & 13169.06  & 20182 & 17214  $\pm$ 2411.5 \\
    Space Invaders      &  1976 $\pm$   893 & 1883.41   &  1652 & 3697.5 $\pm$ 2876.1 \\
    Zaxxon              &  4977 $\pm$  1235 & 7129.33   &  9173 & 30610  $\pm$ 8169.0 \\
  \hline  
  \end{tabular}
  \label{tab:1}
\end{table*}

In order to evaluate the reinforcement learning agent that was selected from the bandit framework in this setting we compare the results of that agent with DQN agents \citep{mnih-dqn-2015}, as well as with the more recent massively parallelized form of the same framework, Gorila \citep{DBLP:journals/corr/NairSBAFMPSBPLM15}. For this comparison 
we selected 6 Atari games spanning the full range, from ones where the algorithms are much better than humans, to some for which human players are much better. The goal here is not to beat the state of the art achieved by some of the recent algorithms e.g. \citep{DBLP:journals/corr/NachumNXS17} 
but to show that the bandit framework is indeed competitive in this setting. In principle, combining the proposed approach with more recent algorithms than the standard A3C and running the training for a longer number of epochs can improve the results further. Nevertheless, the agent selected by the bandit (for Atari typically a version of A3C enhanced with VIME) does better than DQN and Gorila in all 6 Atari games and it either increases the difference when it already is better than the pro human player or it at least reduces the gap when it is not. However, the exact architecture is not always the same, as the optimal set of hyperparameters varies in the different games. 

\section{Conclusions}

We have introduced a bandit framework that offers a principled way of selecting between different reinforcement learning agent architectures while maximizing rewards during the learning process, by focusing the exploration on the most promising agents even early on. Furthermore, the proposed framework was shown to reliably select the best agent (in terms of future expected rewards) after a finite number of steps. In order to achieve these goals a composite surrogate reward was used in combination to the bandit, comprising both the true environment reward as well as a term inspired by VIME \citep{DBLP:journals/corr/HouthooftCDSTA16}, that captures the certainty of the agents regarding the environment dynamics, for a given amount of environment interactions. Under this setting, UCB proved an effective and robust algorithm for selecting the best agent after a finite amount of training steps. 
Experimental results on several classic locomotion and computer game environments show that the bandit strategies outperform both a single non-optimal agent, as well as uniform alternation between the agents.


\end{document}